\documentclass{article}

\usepackage[preprint]{neurips_2026}

\usepackage[utf8]{inputenc} 
\usepackage[T1]{fontenc}    
\usepackage{hyperref}       
\usepackage{url}            
\usepackage{booktabs}       
\usepackage{amsfonts}       
\usepackage{nicefrac}       
\usepackage{microtype}      
\usepackage{xcolor}         
\usepackage{amsmath}
\usepackage{multirow}
\usepackage{graphicx}
\usepackage{amsthm}        
\usepackage{algorithm}     
\usepackage{algorithmic}   
\usepackage{tabularray}    
\usepackage{makecell}      
\usepackage{longtable}     
\usepackage{supertabular}  
\usepackage{enumitem}      
\usepackage{caption}       
\usepackage{tcolorbox}
\tcbuselibrary{skins, breakable}
\usepackage{adjustbox}

\newtcolorbox{methodbox}[1][]{
  enhanced,
  colback=gray!5,              
  colframe=gray!65!black,      
  coltext=black,               
  coltitle=white,              
  colbacktitle=gray!65!black,  
  boxrule=0.7pt,
  arc=2pt,
  left=5pt,
  right=5pt,
  top=5pt,
  bottom=5pt,
  fonttitle=\bfseries,
  title=#1
}

\newtcolorbox{promptbox}[1][]{
  colback=gray!5,
  colframe=gray!50!black,
  title=#1,
  fonttitle=\bfseries,
  boxrule=0.5pt
}

\title{\textsc{SkillMaster}: Toward Autonomous Skill Mastery in LLM Agents}

 \author{%
    Min Yang$^{1,2}$, Jinghua Piao$^{2,3 ~\textbf{*}}$, Xu Xia$^{2,4}$, 
    Xiaochong Lan$^{3}$, Jiaju Chen$^{2,5}$, Yongshun Gong$^{1~\textbf{*}}$, \\
    \textbf{Yong Li}$^{2,3}$ \thanks{Jinghua Piao, Yong Li and Yongshun Gong are co-corresponding authors.}\\ 
    $^{1}$Shandong University \quad                                             
    $^{2}$Zhongguancun Academy \quad                                             
    $^{3}$Tsinghua University \\                                                 
    $^{4}$Southeast University \quad                                             
    $^{5}$University of Science and Technology of China \\ 
    minyang@mail.sdu.edu.cn \quad   
    {\{Pjh22, lanxc22\}@mail.tsinghua.edu.cn} \\                              
    {s-xx25@bza.edu.cn} \quad   
    cjj01@mail.ustc.edu.cn \\
    ysgong@sdu.edu.cn\quad 
    liyong07@tsinghua.edu.cn
  }

\begin{document}

\maketitle

\begin{abstract}

Skills provide an effective mechanism for improving LLM agents on complex tasks, yet in existing agent frameworks, their creation, refinement, and selection are typically governed by external teachers, hand-designed rules, or auxiliary modules. As a result, skills remain external resources to be invoked, rather than capabilities that agents can develop, adapt, and internalize through experience. To endow LLM agents with autonomous skill mastery, we propose \textsc{SkillMaster}, a training framework that teaches agents to create new skills, refine existing skills, and select accumulated skills during task solving. This capability is achieved through three key designs. First, we train agents through trajectory-informed skill review, teaching agents to propose, update, or retain skills based on evidence from completed episodes. Second, each candidate skill edit is designed to be evaluated by its counterfactual utility on related probe tasks, providing a direct learning signal for training skill-editing decisions. Third, we introduce DualAdv-GRPO, which separately estimates advantages for task-solving actions and skill-editing decisions, stabilizing joint training across task solving and skill management. Experiments on ALFWorld, WebShop, and SearchQA show that \textsc{SkillMaster} achieves the best overall performance among all compared methods. Further analysis reveals a marked shift in agent capability: agents trained with \textsc{SkillMaster} can identify skill failures, refine procedural knowledge from trajectory evidence, and transfer improvements to future tasks with limited skill-bank edits. Overall, \textsc{SkillMaster} moves LLM agents beyond mere skill use toward self-improving agents capable of developing, adapting, and applying their own skill repertoires. \footnotetext{Our code is released at \url{https://github.com/sduyangmin/Skill-Master}.}

\end{abstract}

\section{Introduction}

\begin{figure}[t]
\centering
\includegraphics[width=\textwidth]{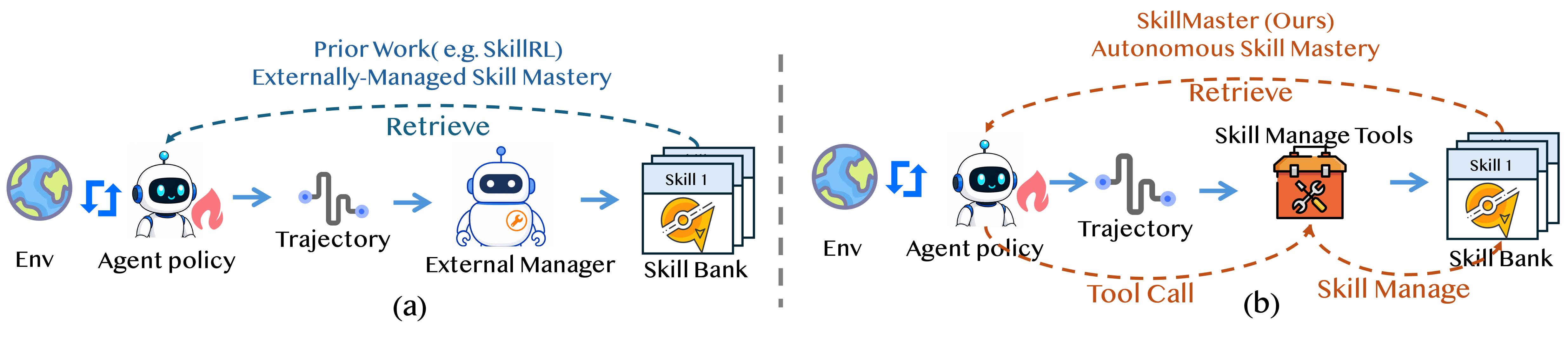}
\caption{\textbf{From externally-managed to autonomous skill mastery.}
  \textbf{(a) Prior work:} Skill management is handled by an external module;
  the agent only retrieves skills. \textbf{(b) \textsc{SkillMaster}:} The agent
  self-manages its skill bank via tool calls, forming a closed loop where skill
  management is a learned RL objective.}
\label{fig:motivation}
\end{figure}
Large language model (LLM) agents have demonstrated impressive capabilities across complex tasks such as embodied household manipulation~\citep{shridharalfworld,wang2023voyager}, web navigation~\citep{yao2022webshop,liu2025wepo,qiu2025evaluating}, information seeking~\citep{jin2025search,zhang2025evolvesearch,li2025search}, and software engineering and code generation~\citep{soni2026coding,rashid2025swe,yu2025utboost,puvvadi2025coding}. Despite strong performance on individual tasks, LLM agents remain largely episodic, failing to effectively leverage past experience for cross-task learning~\citep{xia2026skillrl}. Memory-based methods~\citep{chhikara2025mem0,liu2026simplemem} can store raw
trajectories, but these memories are lengthy and noisy, making it difficult to extract core principles~\citep{xia2026skillrl}.  
To address this limitation, recent work has introduced the concept of skills: compact, reusable abstractions of effective experience that guide future behavior~\citep{xia2026skillrl, wang2026tarse, li2026skillsbench, jiang2026sok, gao2026skillreducer}. 
Unlike raw trajectories, skills distill essential procedures and heuristics, substantially improving task success and efficiency. The practical value is already evident in deployed systems: personal assistant OpenClaw~\citep{openclaw_docs_2026} and coding agent Claude Code~\citep{anthropic_claude_code_2026} both rely on skill-based approaches.
However, existing methods~\citep{xia2026skillrl} typically rely on an external LLM teacher that distills skills from completed trajectories on a fixed schedule, while the main policy only retrieves and applies them. Consequently, skill management remains an external mechanism rather than a learnable component of the agent's policy, limiting autonomous skill mastery.

Recent work has attempted to make skill management learnable~\citep{li2026arise,zhang2026memskill,wu2025cosplay}, 
but these approaches typically rely on auxiliary modules or separate pipelines. 
Moreover, skill management is often guided solely by task outcome rewards, which are sparse and coarse, 
failing to capture how a specific skill edit impacts downstream behavior. 
This lack of explicit skill-quality signals prevents the agent from fully integrating skill management into its own policy.

Consequently, enabling agents to achieve autonomous skill mastery remains challenging. 
\textit{\textbf{First, skills are externally managed, not internally mastered.} } Skill mastery should be an internalized capability of the agent, rather than an external maintenance procedure. Existing training can teach agents to invoke skills, but it rarely teaches them to treat the skill bank as something they can actively improve from their own experience. 
\textit{\textbf{Second, evaluating skill quality is difficult.}} Task success alone is too sparse to indicate whether a specific skill edit helps. Our key insight is that \textit{high-quality skills should produce two measurable downstream effects: increasing success rates on previously failing tasks and reducing steps on already-solvable tasks.} These observable effects provide the explicit signal missing from external management approaches. 
\textit{\textbf{Third, joint optimization is challenging.} }The optimization objectives for skill management and task execution differ, and combining them in a single policy often causes interference, making training unstable.

To address these challenges, we propose the \textsc{SkillMaster} framework, which incorporates skill management into the agent’s learning loop. 
Our framework is built on three key designs. 
The first, \textit{Trajectory-informed skill review}, allows the agent to use tool-integrated reasoning to propose, update, or retain skills based on completed task trajectories, unifying task execution and skill management in an end-to-end reinforcement learning framework. 
The second, \textit{Downstream utility reward}, evaluates each candidate skill modification through counterfactual comparisons on related probe tasks, providing an explicit skill-quality signal for training skill-editing decisions. 
Finally, \textit{DualAdv-GRPO} separately normalizes advantages for task-solving actions and skill-editing decisions, enabling stable joint training of the two optimization objectives within a unified policy.

Our contributions are as follows:
\begin{itemize}[leftmargin=*,itemsep=0pt,topsep=0pt]
\item We propose \textsc{SkillMaster}, a framework that integrates task
    execution with learned skill-management decisions in a single policy,
    jointly optimized through reinforcement learning.
\item We introduce a downstream utility reward that evaluates candidate
    skill revisions by measuring their counterfactual impact on related probe
    tasks.
\item We propose DualAdv-GRPO, which decouples action optimization from skill
  management optimization through separate advantage normalization, enabling
  joint training without objective interference.
\item On ALFWorld, WebShop, and SearchQA, \textsc{SkillMaster}
achieves the best overall performance among all compared methods.
\end{itemize}

\begin{figure}[t]
\centering
\includegraphics[width=\textwidth]{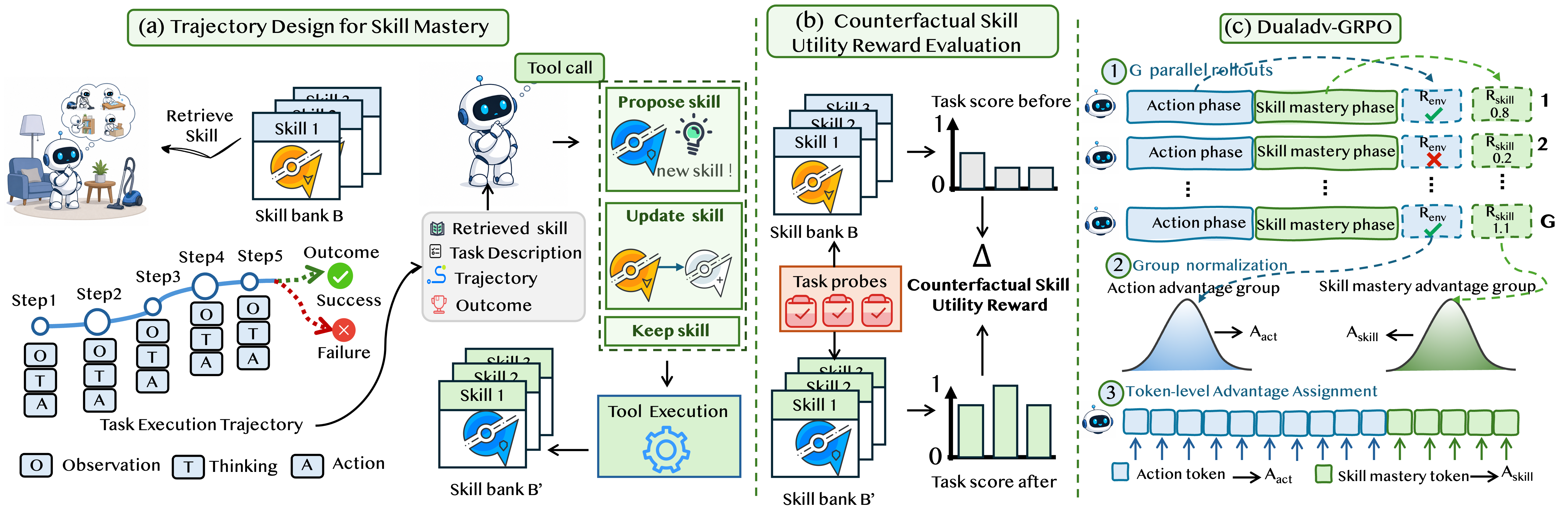}
\caption{\textbf{Overview of \textsc{SkillMaster}.} 
\textbf{(a) Trajectory Design for Skill Mastery:} The agent interacts with the environment guided by retrieved skills and then reviews the episode to propose, update, or retain skills via tool calls. 
\textbf{(b) Counterfactual Skill Utility Reward:} Candidate skill changes are evaluated by counterfactual comparison on related probe tasks. 
\textbf{(c) DualAdv-GRPO:} Action and skill advantages are normalized separately and merged via a tunable weight $\gamma$ into a unified PPO loss for stable joint training.}
\label{fig:overview}
\end{figure}

\section{Method}


Figure~\ref{fig:overview} provides an overview of the \textsc{SkillMaster} framework. 
We describe its three main components, corresponding to the modules illustrated in the figure. 
The first component, \textbf{Trajectory Design for Skill Mastery} (\S\ref{sec:framework}), 
unifies the \emph{Acting Phase} and the \emph{Skill Mastery Phase}: the agent interacts with the environment guided by retrieved skills, and then reviews the episode to propose, update, or retain skills via tool calls. 
The second component, \textbf{Counterfactual Skill Utility Reward} (\S\ref{sec:utility}), evaluates candidate skill changes by comparing performance on related probe tasks using the original and modified skill banks. 
The third component, \textbf{DualAdv-GRPO} (\S\ref{sec:dualadv}) separately normalizes advantages for task-solving actions and skill-editing decisions, merging them via a tunable weight $\gamma$ into a unified PPO loss to stabilize joint training.

\subsection{Trajectory Design for Skill Mastery}
\label{sec:framework}

\textsc{SkillMaster} augments standard agent RL training with an explicit
skill-mastery phase appended after each episode. Every episode proceeds in
two stages:

\begin{itemize}[leftmargin=*,itemsep=1pt,topsep=0pt]
\item \textbf{Acting Phase.} The agent interacts with the environment step by
  step. At each step, relevant skills are retrieved from the skill bank
  $\mathcal{B}$ based on the current task and injected into the observation
  prompt. The agent produces an action and receives a scalar environment reward
  $r_{\text{env}}$. A trajectory $\tau = \{o_0, a_0, r_0, \dots, o_T, a_T, r_T\}$
  is collected.

\item \textbf{Skill-Mastery Phase.} After the episode terminates, the system
  constructs a \emph{skill-review prompt}  that
  presents the task description, the retrieved skills, the trajectory, and the
  final environment feedback. The agent must then output exactly one tool call:
  \texttt{propose\_skill}, \texttt{update\_skill}, or \texttt{keep\_skill}. Specially, we expose skill mastery through three function-calling tools: \texttt{propose\_skill} adds a new skill, \texttt{update\_skill} revises an existing skill, and \texttt{keep\_skill} leaves the bank unchanged. Each call is handled by a skill-management backend. During training, candidate modifications generated by \texttt{propose\_skill} and \texttt{update\_skill} are temporarily applied to the skill bank and evaluated through counterfactual utility evaluation. A modification is committed to the persistent skill bank only when its utility reward satisfies $R_{\text{utility}}>0$; otherwise, the modification is discarded. During evaluation, no utility feedback is available, and all valid skill-management tool calls are directly executed and persisted. After each episode, the skill-review prompt presents the task, outcome, retrieved skills, action-observation trajectory, and final environment feedback. The agent is instructed to reason briefly and output exactly one tool call. Full tool schemas and prompt templates are provided in Appendix~\ref{sec:tool_schemas} and Appendix~\ref{sec:prompt_templates}. 
  This phase receives a dedicated skill-mastery reward $R_{\text{skill}}$
  (defined in \S\ref{sec:utility}).
\end{itemize}

The two phases follow distinct optimization objectives: the acting phase is optimized for immediate environment feedback, whereas the skill-mastery phase is optimized for the long-term quality of the skill bank. We therefore treat them as \emph{heterogeneous phases} and introduce a specialized optimization algorithm to train them jointly (\S\ref{sec:dualadv}).



\subsection{Counterfactual Skill Utility Reward}
\label{sec:utility}

A central challenge in self-managed skill evolution is defining what makes a
skill \emph{good}. In prior work such as \textsc{SkillRL} \citep{xia2026skillrl}, skills are distilled
by an external teacher and the agent is trained with only the environment
outcome reward---a binary success signal that conflates task execution quality
with skill quality. This reward cannot distinguish a genuinely useful skill from
one that merely \emph{looks} plausible, because it provides no signal about
whether the skill actually helps on future tasks.

Our key insight is that a high-quality skill should produce two measurable
effects on tasks that share similar requirements: (1)~tasks that previously
\emph{failed} should become more likely to \emph{succeed}, and (2)~tasks that
already succeeded should be completable in \emph{fewer steps}, since the skill
encodes more efficient strategies. We operationalize this intuition through a
\emph{downstream utility reward} based on counterfactual probe evaluation.

\subsubsection{Probe-Based Counterfactual Evaluation}

When the agent calls \texttt{propose\_skill} or \texttt{update\_skill}, we select
$K$ \emph{probe tasks} that are semantically related to the current episode
(e.g., tasks from the same task family, which share similar skill requirements).
The selection uses a deterministic seed derived from the current task identifier,
ensuring reproducibility. The probe pool is drawn from a held-out set of tasks
from the training, so utility evaluation
measures genuine skill transfer rather than memorization of seen tasks. The
benchmark-specific definitions of probe task are detailed in
\S\ref{sec:setup}.


For each probe task $p_i$, we evaluate the impact of a candidate skill modification by comparing the task under two skill banks:
\[
\mathcal{B} \;\xrightarrow{\;\text{apply candidate mutation}\;}\; \mathcal{B}'.
\]
Specifically, we first rollout $p_i$ using the original skill bank $\mathcal{B}$. 
Next, we create a temporary bank $\mathcal{B}'$ by applying the candidate skill modification, 
and then rollout the same probe task $p_i$ using $\mathcal{B}'$. 
This counterfactual comparison provides a direct measure of the skill modification's effect on task performance.

Each probe rollout is scored according to our two desiderata---success and
efficiency:

\begin{equation}
\text{score}(p_i, \mathcal{B}) = \mathbf{1}[\text{success}_i] \;+\;
\frac{M - \text{steps}_i}{M}
\cdot \mathbf{1}[\text{success}_i]
\label{eq:probe_score}
\end{equation}

where $M$ is the maximum allowed steps. The first term captures whether the
task now succeeds where it may have failed before; the second term captures
step efficiency---a skill that encodes a shorter, more direct strategy yields a
higher score. Failed probes receive $0$, since a skill that does not enable
success confers no benefit regardless of step count.

\subsubsection{Utility Computation}

Define $\delta_i = \text{score}(p_i, \mathcal{B}') - \text{score}(p_i,
\mathcal{B})$ as the per-probe performance delta. Over $K$ probes, let:

\begin{equation}
\bar{\delta} = \frac{1}{K} \sum_{i=1}^{K} \delta_i, \qquad
w = \sum_{i=1}^{K} \mathbf{1}[\delta_i > 0], \qquad
\ell = \sum_{i=1}^{K} \mathbf{1}[\delta_i < 0]
\end{equation}

The utility reward combines the average improvement magnitude with a directional
consistency term:

\begin{equation}
R_{\text{utility}} =  \bar{\delta} + \alpha \cdot \frac{w - \ell}{K},\
\label{eq:utility_reward}
\end{equation}
here, $\alpha$ balances the magnitude of improvement and directional consistency. The directional consistency term $\alpha \cdot (w-\ell)/K$ reduces sensitivity to single-probe outliers, favoring skill edits whose effects are supported by multiple probes. This design encourages skill exploration while ensuring that rewards are assigned only to edits with consistent downstream impact. The full skill-mastery reward is:

\begin{equation}
R_{\text{skill}} = R_{\text{format}} + R_{\text{utility}}
\label{eq:skill_reward}
\end{equation}

where $R_{\text{format}}$ is a composite correctness reward that gives a small
positive bonus for valid tool execution and penalizes parse errors, missing
\texttt{<think>} or \texttt{<tool\_call>} tags, and placeholder content.

\subsection{DualAdv-GRPO: Decoupled Optimization over Heterogeneous Phases}
\label{sec:dualadv}

Standard GRPO normalizes rewards within a group of $G$ rollouts, implicitly
assuming that rewards share a common scale and semantics. This assumption fails
for \textsc{SkillMaster}: acting phases receive binary task rewards
$r_{\text{env}}$, whereas skill-mastery phases receive continuous rewards
$R_{\text{skill}}$ from tool validity and probe utility. Normalizing them
together would entangle task execution with skill evolution and distort
within-type preference ordering.

\subsubsection{Dual-Stream Advantage Estimation}

We address this through \emph{dual-stream advantage normalization}. For each
prompt, we sample $G$ independent trajectories, each containing acting phases
followed by one skill-mastery phase. We compute separate GRPO statistics for
the two reward streams:
$(\mu_{\text{act}}, \sigma_{\text{act}})$ from
$\{r_{\text{env}}^{(j)}\}_{j=1}^{G}$ and
$(\mu_{\text{skill}}, \sigma_{\text{skill}})$ from
$\{R_{\text{skill}}^{(j)}\}_{j=1}^{G}$. The corresponding normalized advantages
are:
\begin{equation}
A_{\text{act}}^{(j)} = \frac{r_{\text{env}}^{(j)} - \mu_{\text{act}}}
                              {\sigma_{\text{act}} + \epsilon},
\qquad
A_{\text{skill}}^{(j)} = \frac{R_{\text{skill}}^{(j)} - \mu_{\text{skill}}}
                                 {\sigma_{\text{skill}} + \epsilon},
\label{eq:dual_adv}
\end{equation}
where $\epsilon$ is a small constant for numerical stability. Thus, advantages assigned to action tokens are normalized only against action
rewards from other trajectories of the same prompt, while advantages assigned to
skill-mastery tokens are normalized only against skill-mastery rewards
from the same prompt group.

\subsubsection{Type-Conditioned Policy Gradient}

We next merge the two normalized advantage streams at the token level. For each
trajectory $j$, let $S_{\text{act}}^{(j)}$ and $S_{\text{skill}}^{(j)}$ denote
the token positions belonging to acting phases and the skill-mastery phase,
respectively. DualAdv-GRPO assigns advantages according to token type:
\begin{equation}
A_{\text{DualAdv}}^{(j)}(l) =
\begin{cases}
A_{\text{act}}^{(j)}, & l \in S_{\text{act}}^{(j)} \\[4pt]
\gamma \cdot A_{\text{skill}}^{(j)}, & l \in S_{\text{skill}}^{(j)}
\end{cases}
\label{eq:merged_adv}
\end{equation}
where $\gamma > 0$ controls the relative weight of the skill-mastery
objective. This preserves within-type preference ordering while allowing action
learning and skill-mastery learning to update a single policy.

The per-token policy gradient is computed through the standard PPO-clipped
objective. Let $r_l(\theta) = \pi_\theta(a_l \mid o_l) / \pi_{\text{old}}(a_l
\mid o_l)$ be the probability ratio and $L$ the number of generated tokens:

\begin{equation}
\mathcal{L}^{(j)}(\theta) = -\frac{1}{L} \sum_{l=1}^{L}
\min\!\Big(
  r_l(\theta) \, A_{\text{DualAdv}}^{(j)}(l),\;
  \operatorname{clip}\!\big(r_l(\theta),\, 1-\varepsilon,\, 1+\varepsilon\big) \,
  A_{\text{DualAdv}}^{(j)}(l)
\Big)
\label{eq:traj_loss}
\end{equation}

The full objective averages over all $N$ trajectories with KL regularization:

\begin{equation}
\mathcal{L}(\theta) = \frac{1}{N} \sum_{j=1}^{N} \mathcal{L}^{(j)}(\theta)
\;+\; \beta \cdot \mathcal{L}_{\text{KL}}(\theta)
\label{eq:full_loss}
\end{equation}

The complete DualAdv-GRPO procedure is summarized in
Algorithm~\ref{alg:dualadv} in Appendix~\ref{sec:algorithm}.

\section{Experiments}

We evaluate  {SkillMaster} on two standard LLM-agent benchmarks: ALFWorld
and WebShop. Our experiments are organized around four research questions:
\begin{enumerate}[leftmargin=20pt,itemsep=0pt,topsep=2pt]
\item[\textbf{Q1.}] Does  {SkillMaster} outperform state-of-the-art methods including
  closed-source LLMs, prompt-based agents, and RL-based approaches?
\item[\textbf{Q2.}] How do individual components affect performance?
\item[\textbf{Q3.}] How does SkillMaster evolve and manage the skill bank??
\item[\textbf{Q4.}] How does  {SkillMaster} manage skills in practice?
\end{enumerate}

\begin{table}[t]
\centering
\caption{\textbf{Performance comparison on ALFWorld and WebShop,} where we report ALFWorld per-family success rates (\%) and overall average success rate, together with WebShop score and success rate. The best results are shown in \textbf{bold} and the second-best results are underlined.}
\label{tab:main_results}
\small
\begin{tabular}{lccccccc|cc}
\toprule
& \multicolumn{7}{c|}{\textbf{ALFWorld}} & \multicolumn{2}{c}{\textbf{WebShop}} \\
\textbf{Method} & \textbf{Pick} & \textbf{Look} & \textbf{Clean} & \textbf{Heat} & \textbf{Cool} & \textbf{Pick2} & \textbf{All} & \textbf{Score} & \textbf{Succ.} \\
\midrule
\multicolumn{10}{c}{\textit{Closed-source LLMs}} \\
\midrule
GPT-4o      & 75.3 & 60.8 & 31.2 & 56.7 & 21.6 & 49.8 & 48.0 & 31.8 & 23.7 \\
Gemini-2.5-Pro & 92.8 & 63.3 & 62.1 & 69.0 & 26.6 & 58.7 & 60.3 & 42.5 & 35.9 \\
\midrule
\multicolumn{10}{c}{\textit{Prompt-based Agentic or Memory-based Methods}} \\
\midrule
Qwen2.5-7B-Instruct  & 33.4 & 21.6 & 19.3 & 6.90 & 2.80 & 3.20 & 14.8 & 26.4 & 7.80 \\
ReAct       & 48.5 & 35.4 & 34.3 & 13.2 & 18.2 & 17.6 & 31.2 & 46.2 & 19.5 \\
Reflexion   & 62.0 & 41.6 & 44.9 & 30.9 & 36.3 & 23.8 & 42.7 & 58.1 & 28.8 \\
Mem0            & 54.0 & 55.0 & 26.9 & 36.4 & 20.8 & 7.69 & 33.6 & 23.9 & 2.00 \\
ExpeL           & 21.0 & 67.0 & 55.0 & 52.0 & 71.0 & 6.00 & 46.3 & 30.9 & 11.2 \\
MemP            & 54.3 & 38.5 & 48.1 & 56.2 & 32.0 & 16.7 & 41.4 & 25.3 & 6.40 \\
SimpleMem       & 64.5 & 33.3 & 20.0 & 12.5 & 33.3 & 3.84 & 29.7 & 33.2 & 8.59 \\
\midrule
\multicolumn{10}{c}{\textit{RL-based Methods}} \\
\midrule
RLOO       & 87.6 & 78.2 & 87.3 & 81.3 & 71.9 & 48.9 & 75.5 & 80.3 & 65.7 \\
GRPO        & 90.8 & 66.1 & 89.3 & 74.7 & 72.5 & 64.7 & 77.6 & 79.3 & 66.1 \\
\midrule
\multicolumn{10}{c}{\textit{Memory-Augmented RL-based Methods}} \\
\midrule
MemRL           & 62.8 & 38.5 & 22.2 & 12.5 & 8.00 & 0.00 & 21.4 & 29.5 & 9.20 \\
EvolveR         & 64.9 & 33.3 & 46.4 & 13.3 & 33.3 & 33.3 & 43.8 & 42.5 & 17.6 \\
Mem0+GRPO       & 78.1 & 54.8 & 56.1 & 31.0 & 65.0 & 26.9 & 54.7 & 58.1 & 37.5 \\
SimpleMem+GRPO  & 89.5 & 36.3 & 60.0 & 50.0 & 64.9 & 26.3 & 62.5 & 67.8 & 46.9 \\
SkillRL & 97.9 & 71.4 & 90.0 & 90.0 & 95.5 & 87.5 & 89.9 & 85.2 & 72.7 \\
\midrule
 {SkillMaster}  & \textbf{100} & \textbf{100} & \textbf{100} & \textbf{97.1} & \textbf{95.7} & \textbf{100} & \textbf{98.7} & \textbf{95.0} & \textbf{82.0} \\
\bottomrule
\end{tabular}
\end{table}

\begin{table}[t]
\centering
\caption{Performance comparison on search-augmented QA
tasks (\%). Baseline results are taken from
\citet{xia2026skillrl}. The best results are shown
in \textbf{bold}, and the second-best results are
\underline{underlined}.}
\label{tab:searchqa}
\small
\setlength{\tabcolsep}{4pt}
\resizebox{\textwidth}{!}{%
\begin{tabular}{lccc|cccc|c}
\toprule
\multirow{2}{*}{\textbf{Method}}
& \multicolumn{3}{c|}{\textbf{Single-Hop QA}}
& \multicolumn{4}{c|}{\textbf{Multi-Hop QA}}
& \multirow{2}{*}{\textbf{Overall}} \\
& \textbf{NQ}
& \textbf{TriviaQA}
& \textbf{PopQA}
& \textbf{HotpotQA}
& \textbf{2Wiki}
& \textbf{MuSiQue}
& \textbf{Bamboogle}
& \\
\midrule
Qwen2.5
& 11.6 & 35.6 & 1.20 & 16.4 & 22.2 & 4.80 & 14.4 & 15.2 \\
CoT
& 12.8 & 35.6 & 3.80 & 16.2 & 22.6 & 6.60 & 24.0 & 17.4 \\
RAG
& 27.4 & 58.2 & 17.8 & 25.8 & 23.2 & 9.40 & 16.8 & 25.5 \\
Search-o1
& 19.4 & 40.6 & 11.4 & 17.0 & 27.0 & 8.60 & 30.4 & 22.1 \\
R1-Instruct
& 21.0 & 44.9 & 17.1 & 20.8 & 27.5 & 6.00 & 19.2 & 22.4 \\
Search-R1
& 39.3 & 61.0 & 39.7 & 37.0 & 40.1 & 14.6 & 36.8 & 38.5 \\
ZeroSearch
& 43.6 & 61.8 & \textbf{51.5} & 34.6
& 35.2 & 18.4 & 27.8 & 39.1 \\
StepSearch
& -- & -- & -- & 38.6
& 36.6 & \textbf{22.6} & 40.0 & -- \\
EvolveR
& 43.5 & \textbf{63.4} & 44.6 & 38.2
& \textbf{42.0} & 15.6 & 54.4 & 43.1 \\
\textsc{SkillRL}
& \underline{45.9}
& \underline{63.3}
& 45.9
& \underline{43.2}
& 40.3
& 20.2
& \textbf{73.8}
& \underline{47.1} \\
\midrule
\textsc{SkillMaster}
& \textbf{49.1}
& 62.9
& \underline{47.0}
& \textbf{45.6}
& \underline{41.6}
& \underline{20.5}
& \underline{70.7}
& \textbf{48.0} \\
\bottomrule
\end{tabular}%
}
\end{table}

\subsection{Experimental Setup}
\label{sec:setup}

\paragraph{Environments.}
We evaluate on ALFWorld~\citep{shridharalfworld}, an embodied household benchmark
spanning six manipulation task families, and WebShop~\citep{yao2022webshop}, an
online shopping benchmark requiring multi-step product search and purchase.
Detailed task descriptions are provided in Appendix~\ref{sec:environments}.

\paragraph{Baselines.}
We compare against four categories of methods: closed-source LLMs as zero-shot agents
(GPT-4o, Gemini-2.5-Pro); prompt-based and memory-augmented frameworks (ReAct, Reflexion,
Mem0, ExpeL, MemP); standard RL methods (RLOO, GRPO); memory-augmented RL approaches
(EvolveR, MemRL, Mem0+GRPO, SimpleMem+GRPO); and SKILLRL as the state-of-the-art
teacher-driven skill evolution baseline. Full baseline descriptions are provided in
Appendix~\ref{sec:baselines}. A detailed
fair comparison statement is provided in Appendix~\ref{sec:fair_comparison}.

\paragraph{Implementation Details.}
We follow the cold-start SFT pipeline of SKILLRL~\citep{xia2026skillrl}, using Claude as a
teacher to generate skill-augmented reasoning traces for the Qwen2.5-7B-Instruct base model.
RL training uses GRPO with group size $G=8$, KL penalty coefficient $0.01$, and learning
rate $1\times 10^{-6}$, on 8$\times$A100 GPUs via the Verl framework with vLLM for rollout
generation. All reported results are averaged over three independent runs. The initial skill bank is adapted from SKILLRL with light deduplication.
Skill-mastery phases are appended after every training episode. The utility reward uses
$K=4$ same-family probes with $\alpha=0.3$, selected by task family for ALFWorld and by
product category for WebShop. Full configuration details are provided in
Appendix~\ref{sec:config}.

\subsection{Main Results}

Table~\ref{tab:main_results} confirms two baseline trends. First, RL-trained agents substantially outperform closed-source LLMs and prompt-based methods, confirming that policy optimization is critical for interactive tasks. Second, structured skill libraries (SkillRL) significantly outperform generic memory augmentation, demonstrating the value of procedural abstraction over raw trajectory storage.

\textsc{SkillMaster} achieves the best performance while removing the external teacher. On ALFWorld, it raises the strongest baseline from 89.9\% to 98.7\%, with perfect success on four of six task families and above 95\% on the remaining two. The average improvement of 8.8 percentage points over SkillRL is particularly notable given that SkillRL already represents a strong starting point near 90\% success. On WebShop, success improves from 72.7\% to 82.0\%, and the overall score increases from 85.2 to 95.0. The gap between score (95.0) and success (82.0) suggests that \textsc{SkillMaster} not only completes more tasks but also achieves higher-quality decisions on tasks it successfully solves. These gains are broadly distributed across all task types, with no family falling below 95\% on ALFWorld, indicating improved general competence rather than overfitting to specific patterns. Table~\ref{tab:searchqa} further reports performance
on seven search-augmented QA datasets covering
single-hop and multi-hop questions.
\textsc{SkillMaster} achieves the highest overall
score of 48.0\%, exceeding \textsc{SkillRL} by
0.9 percentage points.
It improves over \textsc{SkillRL} on five of the
seven datasets and achieves the best performance
on NQ and HotpotQA. Means and standard deviations across repeated runs
on ALFWorld, WebShop, and SearchQA are reported in
Appendix~\ref{sec:run_variability}.

Unlike the teacher-driven approach in SkillRL, \textsc{SkillMaster} learns skill management end-to-end as part of its policy. The downstream utility reward provides a direct quality signal for skill edits via counterfactual comparison on probe tasks, and the agent learns to decide \emph{when} and \emph{how} to revise skills from trajectory evidence. The substantial gains over SkillRL in Table~\ref{tab:main_results} demonstrate the advantage of this utility-guided design over fixed external curation. A detailed case study appears in Section~\ref{sec:Case Study}.

\begin{figure}[t]
\centering
\includegraphics[width=\textwidth]{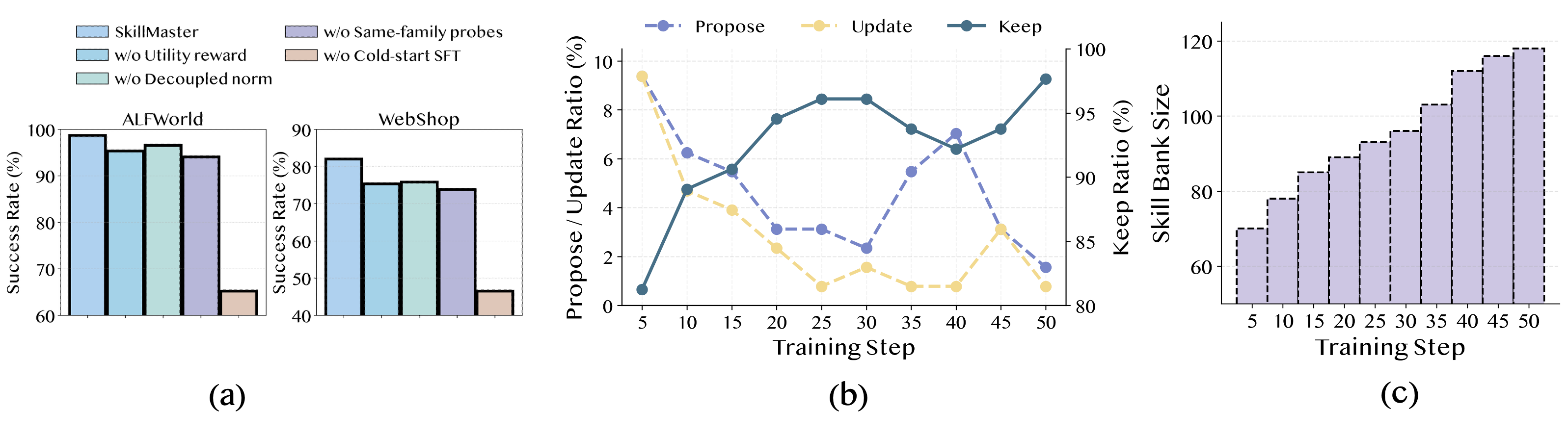}

\caption{\textbf{(a)} Ablation of skill mastery components on ALFWorld and
WebShop. \textbf{(b)} Evolution of skill-management behaviors during ALFWorld
RL training, showing the changing proportions of
\texttt{keep\_skill}, \texttt{propose\_skill}, and \texttt{update\_skill}.
\textbf{(c)} Growth of the skill bank size during ALFWorld RL training.
}
\label{fig:skill_evolution}
\end{figure}

\begin{figure}[t]
\centering
\includegraphics[width=\textwidth]{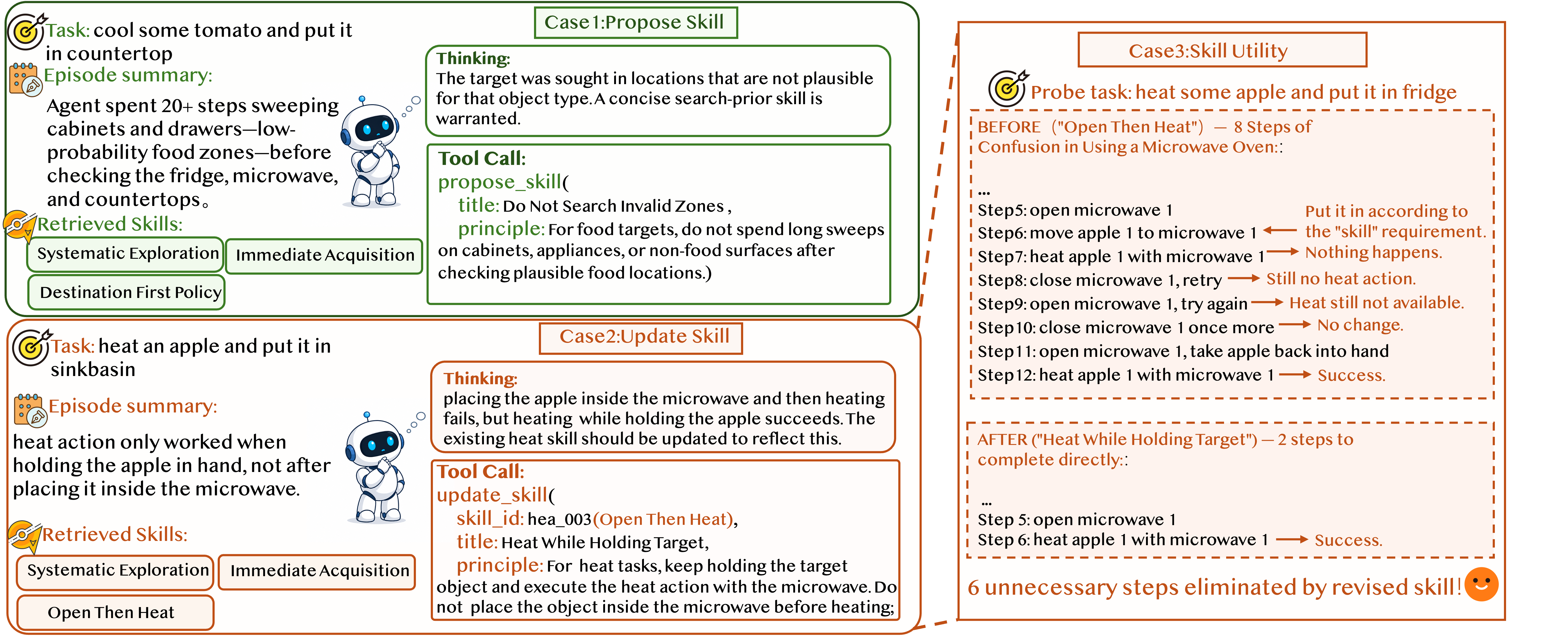}
\caption{\textbf{Case studies of skill management and utility evaluation.}
  \textbf{Case 1 (Propose Skill):} The agent failed a cooling task after
  exhaustively searching low-probability food zones (cabinets, drawers). It
  proposed \texttt{Do Not Search Invalid Zones}, encoding a reusable
  search-prior principle. \textbf{Case 2 (Update Skill):} The agent observed
  that heating succeeded only when holding the apple, contradicting the existing
  skill \texttt{Open Then Heat} (which instructed placing the object inside
  first). It revised the skill to \texttt{Heat While Holding Target}.
  \textbf{Case 3 (Skill Utility):} A probe task (\texttt{heat apple} $\to$
  \texttt{fridge}) was evaluated before and after the skill revision of Case~2.
  Before: the old skill caused an 8-step microwave confusion (place inside
  $\to$ heat fails $\to$ repeated open/close $\to$ take back $\to$ re-heat).
  After: the revised skill enabled correct 2-step execution (open microwave,
  heat while holding), eliminating 6 unnecessary steps.}
\label{fig:case_study}
\end{figure}

\subsection{Ablation Study}
Figure~\ref{fig:skill_evolution}(a) reports the impact of removing each
component of \textsc{SkillMaster}. Removing the utility reward leads to a clear
drop on both benchmarks, suggesting that counterfactual probe evaluation provides
an important signal for distinguishing useful skill edits from merely well-formed
ones. Replacing per-type advantage normalization with a single coupled
normalization group also reduces performance, showing that decoupling action and
skill-mastery  advantages is important for joint optimization. Random probes
consistently underperform same-family probes, showing that task-related probe
selection is important for informative skill-quality feedback. Removing the
cold-start SFT phase causes the largest degradation, highlighting the importance
of initializing the agent with basic tool-use and task-execution capabilities.
Overall, each component contributes to stable and effective self-managed skill
evolution. To further disentangle the effect of utility-credited skill evolution from extra post-episode computation, and to test robustness to weaker initial skill banks, we provide additional control experiments in Appendix~\ref{sec:control_review_compute} and Appendix~\ref{sec:weak_bank}.

\subsection{Analysis of Skill Evolution}
\label{sec:skill_evolution}
We further analyze the skill evolution process on ALFWorld during the RL
training stage, where the agent performs skill-management operations after
each episode. Beyond downstream
task performance, we examine whether \textsc{SkillMaster} can autonomously
create, refine, and retain skills as the skill bank evolves.
Figure~\ref{fig:skill_evolution}(b) shows the evolution of
skill-management decisions, including \texttt{keep\_skill},
\texttt{propose\_skill}, and \texttt{update\_skill}, during ALFWorld training.
Early in training, the agent performs more active skill modifications, with
\texttt{propose\_skill} and \texttt{update\_skill} accounting for 18.76\% of
decisions. As training progresses, the agent gradually shifts toward
skill retention, with \texttt{keep\_skill} increasing from 81.25\% to 97.66\%
while modification decisions decrease to 2.34\%. This transition
indicates that \textsc{SkillMaster} learns to explore new skills initially and
then consolidate useful knowledge as the skill bank matures.
Figure~\ref{fig:skill_evolution}(c) further shows the growth of the skill bank
during ALFWorld training. Starting from 58 initial skills, the bank expands to
118 skills after training, demonstrating that the learned skill-management
policy can continuously discover and preserve new procedural knowledge.
Frozen-checkpoint evaluations further show that
\textsc{SkillMaster} maintains strong performance
after restoring the original initial bank and can
construct useful skills from an empty bank at test time.
Detailed results are provided in
Appendix~\ref{sec:test_time_skill_management}.

\subsection{Case Study}
\label{sec:Case Study}
Figure~\ref{fig:case_study} presents three cases drawn from ALFWorld
rollouts, illustrating how  {SkillMaster} manages its skill bank through TIR.
Full case descriptions are provided in Appendix~\ref{sec:case_study_details}.

Case~1 demonstrates \emph{proposing} a new skill when a gap is detected. The agent
failed a cooling task because it swept low-probability food zones instead of
checking high-probability locations first. During skill review, it recognized
that none of the retrieved skills addressed search-prior allocation, and called
\texttt{propose\_skill} to add a reusable search heuristic.

Case~2 demonstrates \emph{updating} an existing skill when it proves imprecise.
The agent succeeded at a heating task but observed a recurring pattern: the
existing skill instructed placing the object inside the microwave before heating,
yet heating only worked when the object was held in hand. The agent called
\texttt{update\_skill} to correct this operational guidance.

Case~3 \emph{validates} the downstream impact of the revision from Case~2. A
probe task from the same family was evaluated before and after the skill change.
Before the revision, the imprecise skill caused an 8-step confusion at the
microwave. After the revision, the same task completed heating in 2 steps,
eliminating six unnecessary operations. This direct efficiency gain is precisely
what the utility reward is designed to measure and reinforce.

\subsection{Computational Overhead}
\label{sec:skillmanage}
We also analyze computational overhead in Appendix~\ref{sec:overhead}. On
ALFWorld, enabling utility-based skill management increases the average training
time from 12.13 minutes per step to about 16.00 minutes per step, corresponding
to a 31.9\% wall-clock increase. It also raises average GPU memory allocation
from 33.51 GiB to 34.57 GiB and peak GPU memory allocation from 61.29 GiB to
70.49 GiB. These results show that the utility reward introduces noticeable but
manageable computational overhead.

\section{Related Work}

We summarize the most relevant work here and defer a broader discussion to
Appendix~\ref{sec:related_work}.
\paragraph{Skill management in LLM agents.}
Prior agent-memory methods store interaction traces for later retrieval
~\citep{park2023generative,shinn2023reflexion,chhikara2025mem0}, whereas
skill-based methods abstract experience into reusable procedural knowledge
~\citep{wang2023voyager,xia2026skillrl}. In most such approaches, however,
skill management is handled by an external teacher, fixed update rule, or
specialized module rather than by the acting policy itself. Recent concurrent
methods make skill evolution more adaptive but still rely on managerial roles,
controllers, or skill-generation pipelines~\citep{li2026arise,
zhang2026memskill,wu2025cosplay,wang2025reinforcement,zhang2026coevoskills}.
Moreover, they typically assess skill quality only indirectly through task
outcomes or heuristic update rules, rather than explicitly measuring the causal
utility of a particular skill edit. It also leaves skill editing weakly
supervised, since task-level rewards need not reveal whether a specific write
operation was beneficial. \textsc{SkillMaster} instead uses tool-integrated
reasoning to let the same policy both act and edit its own skill bank, with
edits credited by counterfactual downstream utility.
\paragraph{RL and tool-integrated reasoning.}
RL post-training methods such as PPO, RLOO, GRPO, DAPO, and GiGPO optimize LLM
agents through environment feedback~\citep{schulman2017proximal,
ahmadian2024back,shao2024deepseekmath,yu2025dapo,feng2025group}. Tool-use and
TIR methods train models to invoke tools or structured function
calls~\citep{schick2023toolformer,wei2025autotir,fang2026adatir,
zeng2026autotool,zhang2026aster}. Most such methods use tools to access
external information or execute auxiliary computation, rather than to modify
the agent's own reusable memory. Our setting is different because memory-write
actions affect future behavior only indirectly, often across multiple later
episodes. As a result, action decisions and write decisions induce different
credit-assignment structures and should not share a single normalized advantage
signal. Our work extends this paradigm to self-editing skill memory, where
write actions are evaluated by downstream utility and optimized separately from
task-execution turns.



\section{Conclusion and Discussion}

We presented \textsc{SkillMaster}, a framework that enables LLM agents to
manage reusable skill banks through tool-integrated reasoning and reinforcement
learning. The central idea is to turn skill management from an externally
managed maintenance process into an optimization problem within the agent's own
policy, where skill changes are generated through tool calls and credited by
counterfactual downstream utility. This enables the policy to jointly learn how
to act and how to improve the knowledge it will reuse in future episodes.Experiments on ALFWorld and WebShop show that \textsc{SkillMaster}
outperforms standard RL and externally managed skill-library baselines, with
both the utility reward and decoupled optimization contributing to the gains.
These results suggest that learned skill management can be a useful mechanism
for improving the quality of reusable procedural knowledge in LLM agents. At the same time, the current method relies on predefined task groupings for
probe selection, and the cost of probe evaluation grows with episode length and
mutation frequency, even if the overhead is manageable in our current settings.
Moreover, our evaluation is limited to two relatively structured benchmarks,
and further validation on more open-ended environments is needed. Promising
directions for future work include skill deletion, multi-agent skill sharing,
continual learning under distribution shift, and extending TIR to other forms
of agent self-improvement. A more detailed discussion is provided in
Appendix~\ref{sec:discussion}.

\clearpage
\bibliographystyle{unsrtnat}
\bibliography{reference}

\clearpage
\appendix
\section{Skill-Review Prompt Templates}
\label{sec:prompt_templates}

Figures~\ref{fig:alfworld_prompt} and~\ref{fig:webshop_prompt} show the full
skill-review prompt templates used in \textsc{SkillMaster} for ALFWorld and
WebShop, respectively. The prompts are constructed by
\texttt{build\_skill\_management\_prompt} in \texttt{skill\_management.py}.
Placeholders (\texttt{\{task\}}, \texttt{\{category\}}, etc.) are filled
per-episode with the task description, inferred skill category, retrieved
skills, trajectory trace, and outcome.

\begin{figure}[ht]
\centering
\begin{promptbox}[ALFWorld Skill-Review Prompt Template]
\begin{scriptsize}
\begin{verbatim}
You are reviewing a completed ALFWorld episode.
Decide whether the skill bank should be updated based on reusable evidence
from this episode.

Rules:
- Call at most one skill-management tool.
- Choose the tool that best reflects whether the episode reveals a reusable
  lesson that is missing or incorrect in the current bank.
- Use propose_skill only for a genuinely new reusable lesson.
- Use update_skill only when an existing skill should be revised.
- Use keep_skill only when the current retrieved skills already cover the
  observed strategy or failure pattern well enough.
- Base your decision on the task, episode evidence, and the current retrieved
  skills.
- Compare the observed pattern against the retrieved skills explicitly; do not
  choose keep_skill just because the broad task category already has some skills.
- Success alone is not a reason to keep_skill: if a successful episode
  demonstrates a concise reusable strategy that is not already covered,
  propose_skill or update_skill.
- Failure alone is not a reason to change the bank: if the failure does not
  reveal a concrete reusable lesson beyond the current retrieved skills, use
  keep_skill.
- For failed episodes, treat repeated invalid loops, repeated ineffective
  actions after "Nothing happens.", missed visible targets, incorrect subgoal
  switching, and losing track of required object counts as strong evidence for
  propose_skill or update_skill unless an existing retrieved skill already
  states that rule explicitly.
- For successful episodes, prefer propose_skill or update_skill when the
  success depends on a reusable tactic, ordering rule, or search heuristic that
  is not already stated explicitly in the retrieved skills.
- Add or revise a skill only if it is generic, concise, and useful for future
  ALFWorld episodes.
- Do not propose or update a skill that merely restates an already retrieved
  skill with minor wording changes.
- Only write skills grounded in the current ALFWorld household environment,
  task, objects, and receptacles.
- Do not store task-instance details such as specific object instances, room
  names, receptacle IDs, or one-off episode narration unless the lesson is
  clearly reusable.
- Do not include meta-commentary about skill-bank decisions, prompt quality,
  guidance quality, success/failure labels, or whether the bank should change.
- Do not output placeholders, ellipses, half-finished text, or copied
  trajectory fragments.
- Do not output an ALFWorld <action>; the episode is already over.
- Do not choose navigation or environment actions such as look, go to, take,
  move, clean, heat, cool, or done.
- First reason inside <think> </think>, then output exactly one skill-management
  tool call in JSON.
- Keep <think> extremely short: 1-3 sentences, no bullet points, no long
  episode recap, and no copied trajectory details.
- The JSON must be enclosed in <tool_call> </tool_call> tags.
- For task-specific skills, set category to the Skill Category shown below.
- Use category="general" only if clearly reusable across multiple task types.

Output requirements:
- Reason inside <think>...</think>, but keep it brief and decision-focused.
- Final content must be exactly one skill-management tool call wrapped in
  <tool_call>...</tool_call>.
- Inside <tool_call>...</tool_call>, output only valid JSON.
- Do not use markdown code fences.

Task: {task}
Skill Category: {category}
Episode Outcome: {outcome}
Episode Reward: {episode_reward}

Retrieved Skills:
{retrieved_skills_text}

Episode Trajectory:
{episode_trace}
END EPISODE TRAJECTORY.

Now output only the final skill-management decision using the required
<tool_call> JSON format.
\end{verbatim}
\end{scriptsize}
\end{promptbox}
\caption{ALFWorld skill-review prompt template.}
\label{fig:alfworld_prompt}
\end{figure}

\begin{figure}[ht]
\centering
\begin{promptbox}[WebShop Skill-Review Prompt Template]
\begin{scriptsize}
\begin{verbatim}
You are reviewing a completed WebShop episode.
Decide whether the skill bank should be updated based on reusable evidence
from this shopping episode.

Rules:
- Call at most one skill-management tool.
- Use propose_skill only for a genuinely new reusable lesson.
- Use update_skill only when an existing skill should be revised.
- Use keep_skill only when the current retrieved skills already cover the
  observed strategy or failure pattern well enough.
- Every decision must be justified by concrete evidence from THIS EPISODE, not
  by generic real-world shopping advice.
- Compare the observed pattern against the retrieved skills explicitly.
- Success alone is not a reason to keep_skill.
- Failure alone is not a reason to change the bank.
- Favor lessons about search query formulation, attribute filtering, option
  selection, product comparison, and the search-click-option-buy workflow.
- Treat missed required attributes, wrong option selection, premature buying,
  ineffective repeated searches, and losing track of constraints as strong
  evidence for propose_skill or update_skill.
- For keep_skill, the reason must tie to the actual trajectory.
- Do not default to "verify more attributes" unless trajectory evidence
  specifically supports it over a search or result-screening lesson.
- Add or revise a skill only if it is generic, concise, and useful for future
  WebShop tasks.
- Do not store one-off product titles, product IDs, page indices, exact prices,
  or episode narration unless the lesson is clearly reusable.
- Do not output a WebShop <action>; the episode is already over.
- Do not choose environment actions such as search[...], click[...], buy, etc.
- First reason inside <think> </think>, then output exactly one skill-management
  tool call in JSON.
- Keep <think> extremely short: 1-3 sentences.
- The JSON must be enclosed in <tool_call> </tool_call> tags.
- For task-specific skills, set category to the WebShop Category shown below.
- Use category="general" only if clearly reusable across multiple categories.

Shopping Task: {task}
WebShop Category: {category}
Episode Outcome: {outcome}
Episode Reward: {episode_reward}
Goal Index: {goal_idx}
Goal Text: {goal}

Retrieved Skills:
{retrieved_skills_text}

Episode Trajectory:
{episode_trace}
END EPISODE TRAJECTORY.

Now output only the final skill-management decision using the required
<tool_call> JSON format.
\end{verbatim}
\end{scriptsize}
\end{promptbox}
\caption{WebShop skill-review prompt template.}
\label{fig:webshop_prompt}
\end{figure}

\section{DualAdv-GRPO Algorithm}
\label{sec:algorithm}

Algorithm~\ref{alg:dualadv} summarizes one training iteration of the
DualAdv-GRPO optimization loop.

\begin{algorithm}[ht]
\caption{DualAdv-GRPO: One Training Iteration}
\label{alg:dualadv}
\begin{algorithmic}[1]
\REQUIRE Policy $\pi_\theta$, reference policy $\pi_{\text{ref}}$, skill bank
$\mathcal{B}$, environment $\mathcal{E}$, group size $G$, skill weight $\gamma$

\FOR{each prompt $p$ in batch}
  \FOR{$g = 1$ to $G$}
    \STATE Sample trajectory $\tau_g$ from $\pi_\theta$: act in $\mathcal{E}$
          with skills from $\mathcal{B}$, then output one skill-management
          tool call
    \STATE Collect $r_{\text{env}}^{(g)}$ and $R_{\text{skill}}^{(g)}$
  \ENDFOR
  \STATE Compute $A_{\text{act}}^{(g)}, A_{\text{skill}}^{(g)}$ via
        Eq.~\ref{eq:dual_adv} using per-type group statistics
  \FOR{each token $l$ of $\tau_g$}
    \IF{$l \in S_{\text{act}}^{(g)}$}
      \STATE Use advantage $A_{\text{act}}^{(g)}$
    \ELSE
      \STATE Use advantage $A_{\text{skill}}^{(g)}$ weighted by $\gamma$
    \ENDIF
  \ENDFOR
\ENDFOR

\STATE $\mathcal{L} \leftarrow \frac{1}{N} \sum_{j=1}^{N}
        \text{Eq.~\ref{eq:traj_loss}} + \beta \cdot \mathcal{L}_{\text{KL}}$
\STATE $\theta \leftarrow \theta - \eta \nabla_\theta \mathcal{L}$
\RETURN Updated policy $\pi_\theta$
\end{algorithmic}
\end{algorithm}

\section{Implementation and Configuration Details}
\label{sec:config}

\paragraph{Probe task selection.}
For ALFWorld, probe tasks are selected from the same task family as the current episode,
which forms a natural skill-sharing boundary: a skill about microwave usage should affect
all heat tasks but is unlikely to affect clean or pick-and-place tasks. For WebShop, probes
are drawn from the same product category (apparel, footwear, electronics, accessories,
home decor, beauty, health), under the heuristic that skills for searching and filtering
shirts transfer to other apparel tasks while being largely irrelevant to electronics or
footwear. The selection uses a deterministic seed derived from the current task identifier
to ensure reproducibility while avoiding overlap with the current task. Probe rollouts use
a separate environment pool with a distinct seed offset by 50,000 to avoid data leakage.
SearchQA does not provide predefined task families.
We therefore construct an offline semantic probe index
over the training questions using
Qwen3-Embedding-0.6B. For each training question, we
select the $K$ most similar questions according to cosine
similarity, excluding the question itself, and store their
identifiers in advance. During RL training, probes are
retrieved through direct table lookup, without online
embedding computation or search. Across all three
benchmarks, probe candidates are drawn exclusively from
the training split; validation and test examples are never
used for probe selection or utility evaluation.

\paragraph{Training hyperparameters.}
All experiments use GRPO with group size $G=8$, a KL penalty coefficient of $0.01$, and a
constant learning rate of $1 \times 10^{-6}$. The utility reward uses $K=4$ same-family
probes and $\alpha=0.3$. Skill-management turns are configured at trajectory level. The
skill bank uses the standard SKILLRL JSON format with general skills and task-specific
skills organized by category.

\section{Baseline Descriptions}
\label{sec:baselines}

Our baselines span four categories. First, we include strong closed-source LLMs as
zero-shot agents: GPT-4o and Gemini-2.5-Pro serve as upper bounds for general-purpose
reasoning without task-specific training. Second, we compare against prompt-based and
memory-augmented agent frameworks: ReAct and Reflexion use in-context reasoning chains
to improve multi-step decision making; Mem0, ExpeL, and MemP maintain external memory
stores or experience pools to guide agent behavior, but do not update model parameters
through reinforcement learning. Third, we evaluate standard RL-based methods including
RLOO and GRPO, which optimize agent policies through group-relative advantage estimation
over trajectory rollouts. Fourth, we consider memory-augmented RL approaches that embed
persistent memory mechanisms directly into the RL optimization loop: EvolveR, MemRL,
Mem0 combined with GRPO, and SimpleMem combined with GRPO. Finally, we include SKILLRL
as a direct point of comparison, since it represents the state of the art in
teacher-driven skill evolution and shares the same skill bank infrastructure.

\section{Tool Schemas}
\label{sec:tool_schemas}

The three skill-management tools are registered in the multi-turn rollout
infrastructure as standard OpenAI-compatible function-calling schemas.
Figures~\ref{fig:propose_schema}, \ref{fig:update_schema}, and
\ref{fig:keep_schema} show the exact JSON definitions passed to the LLM during
the skill-management turn.

\begin{figure}[ht]
\centering
\begin{promptbox}[propose\_skill]
\begin{scriptsize}
\begin{verbatim}
{
  "type": "function",
  "function": {
    "name": "propose_skill",
    "description": "Propose and add a new reusable skill to the skill bank
                    after a completed episode.",
    "parameters": {
      "type": "object",
      "properties": {
        "category":      {"type": "string",
                          "description": "Skill category."},
        "title":         {"type": "string",
                          "description": "Short skill title."},
        "principle":     {"type": "string",
                          "description": "Reusable principle."},
        "when_to_apply": {"type": "string",
                          "description": "Trigger condition."},
        "evidence":      {"type": "string",
                          "description": "Episode evidence."}
      },
      "required": ["category", "title", "principle",
                   "when_to_apply", "evidence"]
    }
  }
}
\end{verbatim}
\end{scriptsize}
\end{promptbox}
\caption{\texttt{propose\_skill} tool schema.}
\label{fig:propose_schema}
\end{figure}

\begin{figure}[ht]
\centering
\begin{promptbox}[update\_skill]
\begin{scriptsize}
\begin{verbatim}
{
  "type": "function",
  "function": {
    "name": "update_skill",
    "description": "Update an existing skill in the skill bank after a
                    completed episode.",
    "parameters": {
      "type": "object",
      "properties": {
        "skill_id":      {"type": "string",
                          "description": "Existing skill_id or exact
                                          retrieved skill title."},
        "title":         {"type": "string",
                          "description": "Updated title."},
        "principle":     {"type": "string",
                          "description": "Updated principle."},
        "when_to_apply": {"type": "string",
                          "description": "Updated trigger condition."},
        "reason":        {"type": "string",
                          "description": "Why the skill should be revised."}
      },
      "required": ["skill_id", "title", "principle",
                   "when_to_apply", "reason"]
    }
  }
}
\end{verbatim}
\end{scriptsize}
\end{promptbox}
\caption{\texttt{update\_skill} tool schema.}
\label{fig:update_schema}
\end{figure}

\begin{figure}[ht]
\centering
\begin{promptbox}[keep\_skill]
\begin{scriptsize}
\begin{verbatim}
{
  "type": "function",
  "function": {
    "name": "keep_skill",
    "description": "Keep the skill bank unchanged after a completed episode.",
    "parameters": {
      "type": "object",
      "properties": {
        "reason": {"type": "string",
                   "description": "Why no new skill or update is needed."}
      },
      "required": ["reason"]
    }
  }
}
\end{verbatim}
\end{scriptsize}
\end{promptbox}
\caption{\texttt{keep\_skill} tool schema.}
\label{fig:keep_schema}
\end{figure}

\section{Extended Discussion}
\label{sec:discussion}

\paragraph{Probe selection.}
The current probe selection uses task families for ALFWorld and product categories
for WebShop. In more open-ended environments without natural task groupings, an
adaptive similarity metric, such as one based on skill-retrieval embeddings or
learned task representations, would be required. This is a common challenge for any
method that relies on measuring transfer between tasks, and we expect that advances
in task representation learning will directly benefit this aspect of SkillMaster.

\paragraph{Probe evaluation cost.}
Each skill mutation triggers O(K) additional probe rollouts. While the overhead is
modest in our settings, it would grow for environments with longer episodes or for
skill banks that trigger frequent mutations. Strategies such as importance sampling
over probes, amortized evaluation across multiple candidate mutations, or learned
value functions that approximate probe scores could reduce this cost in future work.

\paragraph{Domain scope.}
Our experiments are conducted on two established but relatively structured
benchmarks. Whether SkillMaster transfers to more open-ended environments, such as
software engineering or general web browsing, remains to be validated. The core
mechanism is domain-agnostic: any environment that admits a notion of related tasks
can in principle support counterfactual probe evaluation.

\paragraph{Future directions.}
Beyond addressing the limitations above, several research directions are promising.
First, the utility reward framework could support skill deletion, enabling the agent
to identify and remove outdated or harmful skills rather than only adding and
revising them. Second, multi-agent skill sharing, where agents trained on different
task distributions exchange skills through a shared bank, could accelerate collective
learning. Third, combining SkillMaster with continual learning setups where the task
distribution shifts over time would test whether self-managed skill evolution enables
faster adaptation compared to static baselines. Finally, extending TIR to other forms
of agent self-improvement, such as prompt refinement or tool composition, could
broaden the impact of this approach.

\section{Environment Details}
\label{sec:environments}

ALFWorld is an embodied household benchmark where the agent navigates and manipulates
objects in simulated indoor scenes to complete goal-directed tasks. It contains six
task families: Pick and Place requires retrieving and delivering objects to specified
locations; Look at Object involves examining items under a desklamp; Clean requires
washing objects at a sink; Heat requires warming objects using a microwave; Cool
requires chilling objects in a fridge; and Pick Two requires collecting and delivering
two objects. WebShop is an online shopping benchmark where the agent must search for
products, navigate result pages, select appropriate options such as size and color,
and complete purchases to match a given goal specification.

\section{Extended Case Study Details}
\label{sec:case_study_details}

All cases are drawn from ALFWorld validation rollouts.

\paragraph{Case 1: Propose Skill.}
The agent was tasked with cooling a tomato and placing it on a countertop. It spent
over twenty steps exhaustively sweeping cabinets and drawers, which are
low-probability zones for food targets, before finally checking the fridge and
countertops where the tomato was most likely to be found. The episode failed. During
the subsequent skill-review phase, the agent examined the retrieved skills (Systematic
Exploration, Immediate Acquisition, Destination First Policy) and recognized a gap: none
of them addressed the allocation of search effort across food versus non-food zones.
It called propose\_skill to add Do Not Search Invalid Zones, which instructs future
agents to prioritize high-probability food locations over indiscriminate sweeping.

\paragraph{Case 2: Update Skill.}
The agent was tasked with heating an apple and placing it in a sink basin. The episode
succeeded, but the trajectory revealed an operational imprecision in the existing skill
Open Then Heat, which instructed the agent to open the microwave, place the apple
inside, and then heat it. In practice, heating failed every time the apple was placed
inside first, and only succeeded when the agent held the apple in hand during the heat
action. The agent called update\_skill to revise this skill to Heat While Holding
Target, correcting the operational guidance to keep holding the target object while
executing the heat action.

\paragraph{Case 3: Skill Utility.}
The probe task belongs to the same task family as the episode from Case 2 and was
selected as one of the same-family probes during the utility reward evaluation triggered
by that update. This probe task requires the agent to heat an apple and place it in a
fridge. Before the revision, the skill Open Then Heat caused an 8-step confusion at the
microwave: open the door, place the apple inside, attempt to heat with no effect, close
and reopen the door repeatedly, take the apple back into hand, and finally heat
successfully. After the revision to Heat While Holding Target, the same probe task
completed the heating phase in just 2 steps: open the microwave and heat while holding
the apple. Six unnecessary steps were eliminated purely by correcting the operational
guidance.

\section{Extended Related Work}
\label{sec:related_work}

\subsection{Skill Management in LLM Agents}

Early work on LLM agent memory stores raw interaction histories and retrieves
relevant experiences at inference time. Generative Agents~\citep{park2023generative}
maintain memory streams to support long-term behavior, while
Reflexion~\citep{shinn2023reflexion} uses verbal self-reflection to improve
subsequent attempts. More recent memory-based agents such as
Mem0~\citep{chhikara2025mem0}, MemRL~\citep{zhang2026memrl},
MemP~\citep{fang2025memp}, EvolveR~\citep{wu2025evolver}, and
SimpleMem~\citep{liu2026simplemem} further explore episodic, procedural, or
lifelong memory mechanisms. These methods demonstrate the value of reusing past
experience, but raw or loosely organized memories can become redundant, noisy,
and difficult to maintain as experience accumulates.

A more compact alternative is to abstract trajectories into reusable
\emph{skills}. Voyager~\citep{wang2023voyager} maintains a skill library in
Minecraft by accumulating reusable action programs, enabling open-ended
exploration and reuse of past solutions. {SkillRL}~\citep{xia2026skillrl}
introduces a hierarchical {SkillBank} distilled by a teacher LLM and uses
retrieved skills to guide RL training. Related work on temporal abstraction and
options in reinforcement learning also motivates the use of reusable procedural
units for long-horizon decision making~\citep{sutton1999between}. However, in
these approaches, skill management is typically handled outside the agent's own
learned policy: a teacher model, fixed rule, or predefined procedure decides
what to store and when to update.

Recent concurrent work has begun to make skill management more adaptive.
ARISE~\citep{li2026arise} adopts a hierarchical Manager-Worker architecture
with a shared policy, where the Manager maintains a tiered skill library and
selects relevant skills, while the Worker generates task solutions conditioned
on those skills. MemSkill~\citep{zhang2026memskill} introduces a
Controller-Designer framework, where the Controller selects relevant memory
skills and the Designer revises the skill set based on hard cases.
COS-PLAY~\citep{wu2025cosplay} co-evolves a decision policy with a skill bank
through an agent-managed pipeline that extracts and refines skills from
unlabeled rollouts. Wang et al.~\citep{wang2025reinforcement} use RL to
transform interaction experiences into reusable action-sequence skills, but rely
on fixed extraction procedures rather than learned skill-management decisions.
CoEvoSkills~\citep{zhang2026coevoskills} targets autonomous construction of
multi-file agent skill packages through a Skill Generator and a co-evolving
Surrogate Verifier. These methods move toward adaptive skill evolution, but
they still realize skill management through dedicated managerial roles,
controllers, verifiers, or skill-evolution pipelines, rather than exposing
skill editing as explicit self-directed decisions of the acting policy.

\textsc{SkillMaster} differs from these approaches in two ways. First, it does
not delegate skill evolution to a dedicated managerial role, controller, or
external skill-evolution pipeline. Instead, it applies tool-integrated
reasoning to skill management: the same policy that acts in the environment
also decides, through structured tool calls, whether to keep, propose, or
update skills. This turns skill management into explicit self-editing behavior
within the acting policy itself. Second, the value of a skill edit is not
judged by surface form, heuristic triggers, or task outcome rewards alone.
Instead, each candidate edit is evaluated by its counterfactual downstream
utility on related probe tasks, providing an explicit skill-quality signal for
RL optimization.

\subsection{RL and Tool-Integrated Reasoning}

RL has become a standard post-training paradigm for LLMs and LLM agents.
PPO~\citep{schulman2017proximal} is widely used for policy optimization, and
recent variants such as RLOO~\citep{ahmadian2024back},
GRPO~\citep{shao2024deepseekmath}, DAPO~\citep{yu2025dapo}, and
GiGPO~\citep{feng2025group} further improve stability, scalability, or
group-based credit assignment. These methods have been applied to interactive
settings such as embodied environments, web interaction, software engineering,
and search-based reasoning~\citep{shridharalfworld,yao2022webshop,
yang2024swe,jin2025search}. Our work builds on this RL post-training paradigm
but introduces a heterogeneous trajectory structure, where task-execution turns
and skill-management turns carry different reward semantics.

Tool-use methods train or prompt LLMs to invoke external functions, APIs, or
tools during reasoning. Toolformer~\citep{schick2023toolformer} shows that
language models can learn to use tools through self-supervision, and
Gorilla~\citep{patil2024gorilla} connects LLMs with large-scale API usage.
Recent tool-integrated reasoning methods further combine tool calls with RL,
including AutoTIR~\citep{wei2025autotir}, AdaTIR~\citep{fang2026adatir},
AutoTool~\citep{zeng2026autotool}, and ASTER~\citep{zhang2026aster}. These
works typically treat tools as external services that return information,
perform computation, or execute actions. In contrast, the tool in
\textsc{SkillMaster} edits the agent's own skill memory. This creates a distinct
credit-assignment problem: the value of a write action can only be assessed by
its downstream effect on future behavior. DualAdv-GRPO addresses this setting
by decoupling advantage normalization for action and skill-management turns
while still optimizing a single policy.

\section{Fair Comparison Protocol}
\label{sec:fair_comparison}

To ensure fair evaluation, all methods in our main results are compared under a
consistent protocol. All RL-based methods, including SkillMaster, SKILLRL, GRPO, RLOO,
and memory-augmented RL baselines, share the same backbone model, Qwen2.5-7B-Instruct. For methods that use a skill bank, the same
initial skill set is provided at the start of training. All methods use the same rollout
budget per episode and are evaluated under identical test-time conditions. For SkillMaster
and SKILLRL, the test-time skill retrieval mechanism uses identical top-K settings and
retrieval mode. Closed-source LLM baselines and prompt-based methods are evaluated
zero-shot under the same task instructions and maximum step limits. Results for SKILLRL,
RLOO, GRPO, ReAct, and Reflexion are reproduced from prior work \citep{xia2026skillrl} under the same
evaluation protocol. Any differences in reported numbers reflect only the algorithmic
contributions of each method, not confounding factors such as model scale, training data,
or inference budget.

\section{Computational Overhead}
\label{sec:overhead}

\noindent
\begin{minipage}[t]{0.50\textwidth}
Table~\ref{tab:overhead} reports the computational overhead introduced by
utility-based skill management on ALFWorld. Without the utility reward,
training takes on average 12.13 minutes per step, while enabling the utility
reward increases the average step time to about 16.00 minutes. This corresponds
to an additional 3.87 minutes per step, or roughly a 31.9\% increase in
wall-clock time. 
\end{minipage}%
\hfill
\begin{minipage}[t]{0.48\textwidth}
\centering

\captionof{table}{Computational overhead of utility-based skill management on ALFWorld.}
\label{tab:overhead}
\small
\begin{tabular}{lcc}
\toprule
\textbf{Metric} & \textbf{Without} & \textbf{With} \\
\midrule
Avg. time / step         & 12.13 min & 16.00 min \\
Avg. GPU memory          & 33.51 GiB & 34.57 GiB \\
Peak GPU memory          & 61.29 GiB & 70.49 GiB \\
\bottomrule
\end{tabular}
\end{minipage}

\noindent
We also compare GPU memory usage over the first four training steps. With the
utility reward enabled, the average GPU memory allocation increases from
33.51 GiB to 34.57 GiB, while the peak allocation rises from 61.29 GiB to
70.49 GiB. 
These results suggest that the main computational cost of
\textsc{SkillMaster} comes from the additional probe-based evaluation required
by the utility reward, which increases runtime noticeably while introducing a
more moderate increase in average memory usage.
Overall, the utility reward introduces a noticeable runtime overhead but a
relatively modest increase in average GPU memory allocation, while providing a
direct learning signal for evaluating whether a candidate skill edit improves
downstream performance.

\section{Analysis of Skill-Management Necessity}
\label{sec:control_review_compute}

\noindent
\begin{minipage}[t]{0.48\textwidth}
Table~\ref{tab:review_control} examines two alternative
explanations for the gains of \textsc{SkillMaster}:
additional post-episode review and cold-start SFT.
We compare GRPO with a review-only variant to assess
the effect of appending a review turn, and introduce
an SFT+GRPO control to assess the additional benefit
of the full skill-management framework beyond a shared
SFT initialization and access to the initial skill bank. 
Adding a post-episode review turn without enabling skill
management improves success 
\end{minipage}%
\hfill
\begin{minipage}[t]{0.50\textwidth}
\centering
\vspace{0pt}
\captionof{table}{Analysis of skill-management necessity.
ALFWorld reports overall success rate (\textsc{All}, \%), and WebShop reports
success rate (\textsc{Succ.}, \%).}
\label{tab:review_control}
\small
\begin{tabular}{lcc}
\toprule
\textbf{Method} & \textbf{ALFWorld} & \textbf{WebShop} \\
\midrule
GRPO                    & 77.6 & 66.1 \\
GRPO + Review-Only      & 78.4 & 66.9 \\
SFT+GRPO             & 92.2 & 75.8 \\
\textsc{SkillMaster}    & 98.7 & 82.0 \\
\bottomrule
\end{tabular}
\end{minipage}

\noindent
by only 0.8 percentage points
on each benchmark, indicating limited gains from review
alone. Full \textsc{SkillMaster} remains substantially
stronger than both GRPO and its review-only variant.
These results suggest that the performance gains cannot
be explained merely by appending a review turn, but
reflect the benefit of learning effective skill-management
decisions under utility-based feedback. 
To further examine whether the gains can be attributed
to cold-start SFT, we introduce an SFT+GRPO baseline that
shares the same SFT checkpoint as \textsc{SkillMaster}. This baseline uses a static skill
bank and omits post-episode skill management,
counterfactual probes, and the skill-utility objective.
Compared with SFT+GRPO, \textsc{SkillMaster} improves
ALFWorld and WebShop success by 6.5 and 6.2 percentage
points, respectively, supporting the additional
contribution of the full skill-management framework
beyond the shared SFT initialization.

\section{Robustness to Weaker Initial Skill Banks}
\label{sec:weak_bank}

\noindent
\begin{minipage}[t]{0.50\textwidth}
Table~\ref{tab:weak_bank} evaluates whether the gains of \textsc{SkillMaster}
depend primarily on inheriting a strong initial skill bank from
\textsc{SkillRL}. We vary the initial skill coverage by retaining only a subset
of the original skills, or removing the initial bank entirely, while keeping
the same initialization for both methods. \textsc{SkillMaster} consistently
outperforms \textsc{SkillRL} across all settings, and the performance gap
remains substantial even when the initial skill bank is heavily weakened or
completely removed.
\end{minipage}%
\hfill
\begin{minipage}[t]{0.47\textwidth}
\centering
\captionof{table}{Robustness to weaker initial skill banks. Results report ALFWorld overall success rate (\textsc{All}, \%).}
\label{tab:weak_bank}
\small
\begin{tabular}{lcc}
\toprule
\textbf{Initial bank} & \textbf{\textsc{SkillRL}} & \textbf{\textsc{SkillMaster}} \\
\midrule
Full (100\%)    & 89.9 & 98.7 \\
Reduced (50\%)  & 84.6 & 94.1 \\
Sparse (25\%)   & 78.8 & 92.5 \\
None (0\%)      & 74.8 & 89.4 \\
\bottomrule
\end{tabular}
\end{minipage}

\noindent
Performance degrades gracefully as the initial skill bank becomes weaker, but
\textsc{SkillMaster} remains clearly stronger than the teacher-driven baseline
under the same initialization, including the 0\% setting. These results suggest
that the gains of \textsc{SkillMaster} cannot be explained solely by inheriting
a strong initial skill bank. Instead, the method continues to benefit from its
learned skill-management and policy adaptation even when the initial skill bank
is substantially weakened or removed.

\section{Test-Time Skill Management}
\label{sec:test_time_skill_management}
To further evaluate the test-time skill-management
capability of \textsc{SkillMaster}, we test whether
the trained policy can manage a fresh initial skill
bank and reconstruct useful skills from an empty bank.
While Section~\ref{sec:weak_bank} varies the initial
skill bank during SFT and RL training, the experiments
here keep the trained policy parameters frozen and
vary the skill bank provided at the start of evaluation
and whether test-time skill management is enabled.

\subsection{Skill Management from a Fresh Initial Bank}
\label{sec:fresh_bank}

\noindent
\begin{minipage}[t]{0.50\textwidth}
Table~\ref{tab:fresh_bank} examines whether
\textsc{SkillMaster} retains its performance after
removing the skill-bank changes accumulated during
training. For each trained checkpoint, we compare
evaluation with the training-evolved bank against
evaluation with the original initial bank restored.
Policy parameters remain frozen, and test-time skill
management is enabled in both settings, allowing
valid proposals and updates to persist across episodes.
No counterfactual probes or utility feedback are used.
We also include SFT+GRPO with a static initial bank
as a reference. Restoring the original initial bank reduces ALFWorld
success by only 0.3 percentage points.
On WebShop, the score decreases by 1.9 points, while success increases
\end{minipage}%
\hfill
\begin{minipage}[t]{0.47\textwidth}
\centering
\captionof{table}{Performance with different skill banks
at the start of evaluation. Results are averaged over
three independent runs.}
\label{tab:fresh_bank}
\small
\setlength{\tabcolsep}{3pt}
\begin{tabular}{@{}lccc@{}}
\toprule
\multirow{2}{*}{\textbf{Setting}}
& \textbf{ALFWorld}
& \multicolumn{2}{c}{\textbf{WebShop}} \\
& \textbf{All} & \textbf{Score} & \textbf{Succ.} \\
\midrule
\shortstack[l]{SFT+GRPO\\Static initial bank}
    & 92.2 & 87.7 & 75.8 \\
\addlinespace
\shortstack[l]{\textsc{SkillMaster}\\Training-evolved bank}
    & 98.7 & 95.0 & 82.0 \\
\addlinespace
\shortstack[l]{\textsc{SkillMaster}\\Fresh initial bank}
    & 98.4 & 93.1 & 84.7 \\
\bottomrule
\end{tabular}
\end{minipage}
\noindent

 by 2.7 percentage points.
Compared with SFT+GRPO, the fresh-bank setting improves
ALFWorld success by 6.2 percentage points, WebShop
score by 5.4 points, and WebShop success by
8.9 percentage points.
These results show that \textsc{SkillMaster} maintains
strong performance even after the skill-bank changes
accumulated during training are discarded.

\subsection{Skill-Bank Dependence and Reconstruction}
\label{sec:bank_reconstruction}

On WebShop, we further examine whether \textsc{SkillMaster}
can construct useful skills from an empty bank and how
its performance depends on access to the original skills.

\noindent
\begin{minipage}[t]{0.55\textwidth}
Table~\ref{tab:bank_reconstruction} compares three
evaluation conditions. In the first, the bank starts
empty and both retrieval and skill management are
disabled, so the policy acts using only the task
instruction, environment observations, and current
episode history. In the second, the bank starts empty
but retrieval and skill management are enabled,
allowing newly generated skills to be stored and
reused in subsequent episodes. In the third, the
policy starts with the complete original initial
bank, with retrieval and skill management enabled.
Enabling reconstruction from an empty bank improves
the WebShop score by 6.10 points and success by
6.25 percentage points.
\end{minipage}%
\hfill
\begin{minipage}[t]{0.43\textwidth}
\centering
\captionof{table}{Skill-bank dependence and reconstruction
on WebShop using the same frozen checkpoint.
Updates indicate whether test-time skill management
is enabled. }
\label{tab:bank_reconstruction}
\small
\setlength{\tabcolsep}{3pt}
\begin{tabular}{@{}lccc@{}}
\toprule
\textbf{Initial bank} & \textbf{Updates}
    & \textbf{Score} & \textbf{Succ.} \\
\midrule
Empty         & Disabled & 72.35 & 65.63 \\
Empty         & Enabled  & 78.45 & 71.88 \\
Full original & Enabled  & 93.55 & 85.16 \\
\bottomrule
\end{tabular}
\end{minipage}

 Since both conditions start
without external skills, this improvement indicates
that the frozen policy can construct useful skill
memory from evaluation trajectories and reuse it
across episodes. However, the reconstruction setting
remains 15.10 score points and 13.28 success-rate
points below the full-original-bank setting, where
test-time skill management is also enabled.
These results show that \textsc{SkillMaster} can
build and use a skill bank without access to the
original skills, but reconstruction only partially
compensates for their removal. Explicit skill memory
therefore remains important for task performance.

\section{Results Across Repeated Runs}
\label{sec:run_variability}

Table~\ref{tab:run_variability} reports the mean and
standard deviation of \textsc{SkillMaster}'s performance
over three runs on ALFWorld, WebShop, and SearchQA.
We include both aggregate results and individual
task-family or dataset results. On ALFWorld, \textsc{SkillMaster} achieves an overall
success rate of $98.7 \pm 0.4$\%, with perfect success
across all three runs on four task families.
On WebShop, the average score and success rate are
$95.0 \pm 1.0$ and $82.0 \pm 0.8$\%, respectively.
On SearchQA, the overall result is $48.0 \pm 0.1$\%.

\begin{table}[t]
\centering
\caption{Performance of \textsc{SkillMaster} across
three runs, reported as mean $\pm$ standard deviation.}
\label{tab:run_variability}
\small
\setlength{\tabcolsep}{2pt}

\resizebox{\textwidth}{!}{%
\begin{tabular}{@{}l*{8}{c}@{}}

\toprule
\multicolumn{9}{c}{\textbf{ALFWorld}} \\
\midrule
\textbf{Metric}
& \textbf{Pick}
& \textbf{Look}
& \textbf{Clean}
& \textbf{Heat}
& \textbf{Cool}
& \textbf{Pick2}
& \multicolumn{2}{c}{\textbf{All}} \\
\midrule
Mean $\pm$ Std.
& $100.0 \pm 0.0$
& $100.0 \pm 0.0$
& $100.0 \pm 0.0$
& $97.1 \pm 0.3$
& $95.7 \pm 1.5$
& $100.0 \pm 0.0$
& \multicolumn{2}{c}{$98.7 \pm 0.4$} \\
\midrule
\multicolumn{9}{c}{\textbf{WebShop}} \\
\midrule
\textbf{Metric}
& \multicolumn{4}{c}{\textbf{Score}}
& \multicolumn{4}{c}{\textbf{Succ.}} \\
\midrule
Mean $\pm$ Std.
& \multicolumn{4}{c}{$95.0 \pm 1.0$}
& \multicolumn{4}{c}{$82.0 \pm 0.8$} \\
\midrule
\multicolumn{9}{c}{\textbf{SearchQA}} \\
\midrule
\textbf{Metric}
& \textbf{NQ}
& \textbf{TriviaQA}
& \textbf{PopQA}
& \textbf{HotpotQA}
& \textbf{2Wiki}
& \textbf{MuSiQue}
& \textbf{Bamboogle}
& \textbf{Overall} \\
\midrule
Mean $\pm$ Std.
& $49.1 \pm 0.3$
& $62.9 \pm 0.1$
& $47.0 \pm 0.1$
& $45.6 \pm 0.1$
& $41.6 \pm 0.7$
& $20.5 \pm 0.2$
& $70.7 \pm 0.2$
& $48.0 \pm 0.1$ \\
\bottomrule
\end{tabular}%
}
\end{table}

\end{document}